# MULTIPLE AUTHORS DETECTION: A QUANTITATIVE ANALYSIS OF DREAM OF THE RED CHAMBER

XIANFENG HU, YANG WANG, AND QIANG WU

Abstract. Inspired by the authorship controversy of *Dream of the Red Chamber* and the application of machine learning in the study of literary stylometry, we develop a rigorous new method for the mathematical analysis of authorship by testing for a so-called chrono-divide in writing styles. Our method incorporates some of the latest advances in the study of authorship attribution, particularly techniques from support vector machines. By introducing the notion of relative frequency as a feature ranking metric our method proves to be highly effective and robust.

Applying our method to the Cheng-Gao version of *Dream of the Red Chamber* has led to convincing if not irrefutable evidence that the first 80 chapters and the last 40 chapters of the book were written by two different authors. Furthermore, our analysis has unexpectedly provided strong support to the hypothesis that Chapter 67 was not the work of Cao Xueqin either.

We have also tested our method to the other three Great Classical Novels in Chinese. As expected no chrono-divides have been found. This provides further evidence of the robustness of our method.

## 1. Introduction

*Dream of the Red Chamber* (红楼梦) by Cao Xueqin (曹雪芹) is one of China's Four Great Classical Novels. For more than one and a half centuries it has been widely acknowledged as the greatest literary masterpiece ever written in the history of Chinese literature. The novel is remarkable for its vividly detailed descriptions of life in the 18th century China during the Qing Dynasty and the psychological affairs of its large cast of characters. There is a vast literature in *Redology*, a term devoted exclusively to the study of *Dream of the Red Chamber*, that touches upon virtually all aspects of the book one can imagine, from the analysis of even minor characters in the book to in-depth literary study of the book. Much of the scope of Redology is outside the focus of this paper.

The original manuscript of *Dream of the Red Chamber* began to circulate in the year 1759. The problems concerning the text and authorship of the novel are extremely complex and have remained very controversial even today, and they remain an important part of Redology studies. Cao, who

The second author is supported in part by the National Science Foundation grant DMS-0813750 and DMS-1043034.

died in 1763-4, did not live to publish his novel. Only hand-copied manuscripts – some 80 chapters – had been circulating. It was not until 1791 the first printed version was published, which was put together by Cheng Weiyuan (程伟元) and Gao E (高鄂) and was known as the *Cheng-Gao version*. The Cheng-Gao version has 120 chapters, 40 more than the various hand-copied versions that were circulating at the time. Cheng and Gao claimed that this "complete version" was based on previously unknown working papers of Cao, which they obtained through different channels. It was these last 40 chapters that were the subject of intense debate and scrutiny. Most modern scholars believe that these 40 chapters were not written by Cao. Many view those late additions as the work of Gao E. Some critics, such as the renowned scholar Hu Shi (胡适 ), called them forgeries perpetrated by Gao, while others believe that Gao was duped into taking someone else's forgery as an original work. There is, however, a minority of critics who view the last 40 chapters as genuine.

The analysis of the authenticity of the last 40 chapters has largely been based on the examination of plots and prose style by Redology scholars and connoisseurs. For example, many scholars consider the plotting and prose of the last 40 chapters to be inferior to the first 80 chapters. Others have argued that the fates of many characters in the end were inconsistent with what earlier chapters have been foreshadowing. A natural question is whether a mathematical stylometry analysis of the book can shed some light on this authenticity debate.

The problem of style quantification and authorship attribution in literature goes at least as far back as 1854 by the English mathematician Augustus De Morgan [7], who in a letter to a clergyman on the subject of Gospel authorship, suggested that the lengths of words might be used to differentiate authors. In 1897 the term *stylometry* was coined by the historian of philosophy, Wincenty Lutaslowski, as a catch-all for a collection of statistical techniques applied to questions of authorship and evolution of style in the literary arts (see e.g. [12]). Today, literary stylometry is a well developed and highly interdisciplinary research area that draws extensively from a number of disciplines such as mathematics and statistics, literature and linguistics, computer science, information theory and others. It is a central area of research in statistical learning (see e.g. [9]). A popular classic technique for stylometric analysis of authorship involves comparing frequencies of the so-called *function words*, a class of words that in general have little content meaning, but instead serve to express grammatical relationships with other words within a sentence. Although this technique is still widely used today, the field of literary stylometry has seen impressive advances in recent years, with more and more new and sophisticated mathematical techniques as well as softwares being developed. We shall not focus on these advances here. Instead we refer all interested readers to the excellent survey articles by Juola [10] and Stamatatos [14] for a comprehensive discussion of the latest advances in the field.

Although there is a vast Redology literature going back over 100 years, the number of studies of the book based on mathematical and statistical techniques is surprisingly small, particularly in view of the fact that such techniques have been used widely in the West for settling authorship questions. Among the notable efforts, Cao [3] meticulously broke down a number of function characters and words into classes according to their functions. By analyzing their frequencies in the first 40 chapters, the middle 40 chapters and the last 40 chapters, Cao concluded that the first 80 chapters and the last 40 chapters were written by different authors. Zhang & Liu [2] examined the occurrence of 240 characters in the book that are outside the GB2312 encoding system, and found that 210 of them have appeared exclusively in the first 80 chapters while only 20 of them have appeared exclusively in the last 40 chapters. This led to the same conclusion by the authors. Yue [1] studied the authorship by combining both historical knowledge and statistical tools. In the study Yue tested two hypotheses, that the last 40 chapters were not written by the same author or they were written by the same author. His study focused on the frequencies of 5 particular function characters, the proportion of texts to poems in each chapter, and a few other stylometric peculiarities such as the number of sentences ended with the character "Ma" (吗). Using various statistical techniques the comparisons led the paper to draw the conclusion that it is unlikely that the first 80 chapters and the last 40 chapters were written by the same author. At the same time, using historic knowledge about the book and the original author Cao Xueqin, the paper also speculated that it was not likely that the last 40 chapters were created entirely by a single different author such as Gao E. In the opposite direction, the studies of Chan [6] and Li & Li [11] concluded that the entire book was likely written by a single author. The study [11] focused on the usage of functional characters while [6] examined the usage of some eighty thousand characters. Both studies tabulated the frequencies of the selected characters, which led to a frequency vector for each of the first 40 chapters, the middle 40 chapters and the last 40 chapters. The correlations of these frequency vectors were computed. In [11] the correlations were found to be large enough for the authors to conclude that the entire 120 chapters of the book were written by the same author. In [6] a fourth frequency vector using parts of the book *The Gallant Ones* (儿女英雄传) was added for comparison. The author found significantly higher correlations among the first three frequency vectors from chapters of *Dream of the Red Chamber* than the correlations between the fourth frequency vector and the first three. This fact formed the basis of the conclusion by the author that all 120 chapters were written by a single author. A different conclusion was reached by Li [4]. By analyzing the frequencies of 47 functional characters and applying several statistical tests the author conjectured that the last 40 chapters were put together by Gao E using unedited and unfinished manuscripts by Cao Xueqin and his family members.

Although some of these aforementioned studies are impressive in their scopes, missing conspicuously from the Redology literature are studies based on the latest advances in literary stylometry, particularly some of the new and powerful methods from machine learning theory. While comparing the frequencies of function characters and words is clearly a viable way to analyze the authorship question, care needs to be taken to account for random fluctuations of these frequencies, especially when some of the function characters and words used for comparison have limited occurrences overall in the book and some times not at all in some chapters. None of the aforementioned studies employed cross validation to address random fluctuations. We have substantial reservations about drawing conclusions from correlations alone as in the studies of Chan [6] and Li & Li [11], because the differentiating power of any single variable such as correlation is rather limited. It would be interesting to see a more comprehensive study of correlations on a large corpus of texts in Chinese to determine its effectiveness as a metric for authorship attribution, something the authors failed to do in both studies. The use of the book *The Gallant Ones* in [6] for benchmark comparison is curious to us in particular, especially considering that the author did not limit to just function characters. The two books are of two different genres and are different in their respective background settings. Considering these differences *and* the fact that *The Gallant Ones* is known *not* to be written by Cao Xueqin, it would be a shock if the correlation between the last 40 chapters of *Dream of the Red Chamber* and the first 80 chapters is *not* higher than the correlation between the last 40 chapters and *The Gallant Ones*. It is possible that the correlation computed in [6] tells more about the genre than the authorship of the books. Again, without extensive evidence that using the same technique the correlation between two bodies of texts written by different authors is generally low even when the plots are closely related, the argument made in [6] is unconvincing at best. The objective of this paper is to present a rigorous stylometric analysis of *Dream of the Red Chamber* that incorporates some of the latest advances in the study of authorship attribution, particularly techniques from the theory of machine learning. To minimize the impact of random fluctuations we have meticulously followed well established protocols in selecting significant features by proper randomization of training and testing samples.

We shall detail our methodology in the next section, including feature construction and selection techniques in machine learning. In Section 3, we use our approach to the study of authorship of *Dream of the Red Chamber* and show the experimental results. In Section 4 we use our approach to analyze the other three Great Classical Novels. Finally in Section 5 we present some additional comments and our conclusions.

## 2. Chrono-divide and Methodology

The main idea behind statistically or computationally-supported authorship attribution is that by measuring some textual features we can distinguish between texts written by different authors. Nearly a thousand different measures including sentence length, word length, word frequencies, character frequencies, and vocabulary richness functions had been proposed thus far [13] over the years. Some of these measures, such as frequencies of function words, have proven effective while others, such as length of words, have proven less effective [10]. The field of literary stylometry has seen impressive advances over the years, and has become an increasingly important research field in the digital age with the explosion of texts online.

This paper focuses on a particular class of authorship controversies, in which there is a suspected change of authorship at some point of a book. In other words, one suspects that the first $X$ chapters of a book were written by one author while the remaining $Y$ chapters were written by another. Clearly, the authorship controversy for *Dream of the Red Chamber* falls into this category. Since no two authors have exactly the same writing style, no matter how similar they might be, a book written in such a fashion would have a stylistic discontinuity going from Chapter $X$ to Chapter $X + 1$. If we can quantify the styles of the two authors by a stylometric function $S(n)$ (a classifier) where $n$ denotes chapters, or chronologically ordered samples, of the book in question, this stylistic discontinuity will appear as a dividing point in the stylometric function $S(n)$ going from $n = X$ to $n = X + 1$. Because the samples are ordered by time, we shall call this divide in the stylometric function $S(n)$ a *chrono-divide in style*, or simply a chrono-divide. This paper develops a technique for verifying and detecting chrono-divides in books or other body of texts. Knowing $X$ and $Y$, as it is the case with *Dream of the Red Chamber*, can help validating the conclusion but is not always necessary for our method. Our method does not apply to any body of texts where two authors share the writing in an interwoven way without a chrono-divide.

The underlying principle of our study is that if a book is in fact written by two authors A and B, then there should exist a group of features that characterize the difference of their respective styles. These features will lead to a stylometric function that separate the book into two different classes. In the rest of the paper we shall use the more conventional term *classifier* for such a stylometric function. The foundational principle for literary stylometry is built around finding such classifiers. Suppose that a chrono-divide in style exists. Then an effective classifier will show a break point somewhere in the middle of the book, before and after which the classifier gives positive values and negative values, respectively. Thus in analyzing a book suspected to be written by two authors with a chrono-divide, one can look for a classifier that gives rise to such a break point. The existence of such a classifier will provide strong support for the two-author hypothesis. Conversely, if such a classifier cannot be found then we can confidently reject the two-author with a chrono-divide hypothesis.

We use function characters and words to build and select a group of stylometric features having the highest discriminative power, and from which we construct our classifier. We shall detail our method in the following subsections.

2.1. **Initial stylometric feature extraction.** Suppose the book in question is suspected to be written by two authors. For simplicity we shall call the part written by author A *Part A* and the part written by author B *Part B*. In many cases, such as with *Dream of the Red Chamber*, both Part A and part B are known. In some cases, they are not precisely known. However, for books suspected to have a chrono-divide from authorship change, there is usually a good estimate for where the divide is. Typically the first few chapters can be confidently attributed to A and the last few chapters to B.

We begin by choosing a feature set consisting of the kinds of features that might be used consistently by a single author over a variety of writings. Typically, these features include the frequencies of words (or characters for books in Chinese), phrases, mean and variation of sentence length, and frequencies of direct speeches and exclamations, and others. In our analysis, to get a better understanding of an author's writing style, we first find the most frequently used characters and words in the book, e.g. we would find the 500 most frequently used characters in the whole book, from which we pick out only, say, $n$ function characters. We choose $m$ words (combinations of characters) among the 300 most frequently used words in the same way. An important point is that by selecting only function characters and words we obtain a selection of characters and words that are *content independent*. This leads to an initial set of features consisting of the frequencies of the $n$ characters and the $m$ words, plus the mean and variance of sentence length as well as the frequencies of direct speeches and exclamations. These features will be computed over given sample texts of the book (e.g. chapters). We normalize each sample text in the following way: set the median of the mean and variation of sentence length and the frequencies of direct speeches, exclamations, $n$ characters and $m$ words in each work of A and B to be 1. For each sample, we now get $n + m + 4$ features.

2.2. **Data preparation.** Having constructed the appropriate feature vectors, we build a distinguishing model through a machine learning algorithm. To do so requires careful data preparation. Since we usually have in hand only limited samples while the number of features will be very large, building a model directly on the entire book will easily lead to over-fitting. To overcome the over-fitting problem, we use the standard technique of separating the whole data into samples consisting of training data and test data. Our model will be established based only on the training data while its performance is tested over the independent test data. If we know Part A and Part B already then a subset of each can be designated as training data. For books suspected to have a chrono-divide in style, the training data will consist of the first few chapters and the last few chapters. The rest of the book will be used as test data.

In order to obtain more training sets and testing sets we shall chunk the book in question into smaller pieces of sample texts of relatively uniform size and style. In all the books we have studied, we have kept the sample texts to be at least 1000 characters long. In the case of *Dream of the Red Chamber* each sample text is a chapter.

2.3. **Feature subset selection.** When we build authorship analysis the model using the training data only, we do not use all the features ($n + m + 4$ features). Instead we start out with all of them, but eventually select a subset of features that achieves the highest discriminative powers. Feature subset selection has been well understood for high dimensional data analysis in the machine learning context. First, the number of discriminative features may be small because the number of features an author uses in a consistently different way from others is usually not very big. Moreover, the classifier can perform very poorly if too many irrelevant features are included into the model. In this paper we will use Support Vector Machines Recursive Feature Elimination (SVM-RFE) introduced in [8] to realize feature selection.

SVM-RFE is a feature ranking method. Given the a set of samples we can use linear SVM to build a linear classifier. It ranks the importance of the features according to their weights. As mentioned above, because of large feature size and small sample size, the classifier might not be robust. In addition, the high correlation between the features may result in small weights for relevant features. Thus the ranking by SVM classifier directly may be inaccurate. In order to refine the ranking, the least important feature is removed and the linear SVM classifier is retrained. This new classifier provides a refined ranking for the remaining features. The process is then repeated until the ranking of all features are refined. This is the SVM-RFE method introduced [8]. The idea underlying SVM-RFE is that in each repeat, although the overall ranking may be poor, the least important feature is seldom a relevant one. By iteratively eliminating the least important features the new classifiers will become more and more reliable and hence will provide better and better ranking. In the application of gene expression data analysis SVM-RFE has been proven to be substantially superior to the SVM direct ranking without RFE.

However in general SVM-RFE is not stable under the perturbation of samples. A small change in samples may result in very different feature ranking. There are two possible reasons. One is that the highly correlated variables are too sensitive and may be ranked in different orders by different classifiers. Another is that, due to the randomness, some subset of samples might be singular in the sense that they are less representative for the whole data structure. As a result the SVM classifiers are over-fitting and the feature ranking by SVM-RFE is therefore unreliable. The first situation is less harmful for classification performance while the second is vital. To overcome this phenomenon and guarantee the stability of the ranking, we use a pseudo-aggregation technique. We randomly choose a subset of training samples to run SVM-RFE to select the top important features. This process is repeated tens or hundreds times and only those features that appear important very frequently are deemed as truly important ones. This removes the randomness and results in a much more reliable ranking.

With this ranking of features, we can conclude which statistics are useful for quantifying the writing style. We use cross validation to select the number of features included in the final classification model. This group of features is a stable and most discriminative subset of features. A final classifier is built to classify the test data.

2.4. **Data analysis.** The classifier we have built is used to analyze the authorship question. We examine the discriminative power of the classifier on the training data. If it cannot even reliably classify the training data we can convincingly reject the two-author hypothesis. Even if it can the telling story will be whether it can classify, or detect a chrono-divide, from the test data. If it fails then again we should reject the two-author hypothesis. On the other hand, if the the classifier classifies the training data, and it can also classify the test data accurately or detect a clear chrono-divide, we can then convincingly conclude that the book does contain two different writing styles and can therefore be confidently attributed to two different authors. Moreover, the feature subset and the classifier describe the difference of the two authors' writing styles.

2.5. **The algorithm.** In the following we summarize the process of our algorithm:

(1) Initialize the data (the book), which contains parts A and B suspected to be written by two different authors.
(2) Split part A and part B into many sections and extract the features for each section as described in section 2.1. This forms the whole data set $D$, containing $D_A$ and $D_B$.
(3) Choose a portion (e.g. 20%-30%) of $D_A$ and $D_B$ respectively to form the test data set and leave the remaining as the training data set. The test data will not be used until the final model is built.
(4) Randomly choose a subset from the training data as modeling data and the rest (again 20%-30%) as the validation data. Run SVM-RFE on the modeling data and using the validation data to determine all the parameters used. This provides a ranking of all the $n+m+4$ features extracted in step 2.
(5) For $d$ range from 1 to $n+m+4$, build a classifier using only the top $d$ features and evaluate their performance on the validation data. The best model is the one with minimal validation error and minimal number of top features. The feature subset of this best model is recorded.
(6) Repeat $T$ times step 4 and step 5 to obtain $T$ best models and $T$ subsets of corresponding important features. We recommend $T$ to be larger than 50. Rank all the features in these subsets according to their appearance frequency. Denote $N$ as the total number of features included.
(7) For $d = 1, ..., N$, using cross validation to select the number of features that should be included in the final classifier. Denote it by $d_*$. Note we require both the cross validation error and the number of features to be as small.
(8) Retrain the model using the whole training set based on this top $d_*$ important features.

(9) Using the classifier to classifying the test data. Draw the conclusion according to the performance.

Since our ranking process involves aggregation of large number of models that are trained using SVM-RFE based on different subsets of the same data source, we refer to our approach as pseudo-aggregation SVM-RFE method.

## 3. Analysis of *Dream of the Red Chamber*

Having established a rigorous protocol for the study of authorship of a body of texts, we apply this protocol to investigate the authorship controversy of the Cheng-Gao version of *Dream of the Red Chamber*. In particular we investigate the existence of a chrono-divide at Chapter 80.

The book is first divided into samples. To balance the number of samples, we generate one sample for each of the first 80 chapters while using the conventional practice of duplicating each of the last 40 chapters into two chapters to obtain 80 samples. From those samples we extract the features by calculating the statistics proposed in subsection 2.1. These features are then normalized for fair comparison. In total we have 196 variables. They are the 144 characters and 48 words, the normalized mean and variation of sentence length, and the frequencies of direct speeches and exclamations.

To investigate the authorship controversy we perform three separate tests. First we build a classifier for the whole book and look for the existence of a chrono-divide at Chapter 80. For added robustness and reliability we also perform the same tests only on the first 80 chapters and the last 40 chapters.

3.1. **Separability of the chapters by Cao and Gao.** In the first experiment we apply our method to the whole Chen-Gao version of *Dream of the Red Chamber*. Samples from the first 60 chapters are designated as training samples for one class while samples from the last 30 chapters are designated as training samples for another class. The remaining samples, from Chapter 61 to 90, are held out as test samples. The training samples are further randomly split into modeling data of 80 samples and validation data of 40 samples. The SVM-RFE is repeated 100 times and $d_*$ is chosen using 50 cross validation runs. We have the following observations.

   ***Instability of SVM-RFE.*** The randomness of the modeling set has resulted in very substantial fluctuations in the number of features selected as well as feature rankings. The resulted classifier may also perform quite differently. Table 3.1 lists the features selected using two different modeling data sets. One selects 11 features and the other selects only 4, with only one feature in common. The classifiers also perform differently. The experiments clearly establish the instability of SVM-REF.

| Modeling set | Features Selected | Validation Error |
|---|---|---|
| 1 | 去，得，就，回，知，到，时，呢，倒，别，作 | 5/40 |
| 2 | 回，方，没，好些 | 1/40 |

Table 1. The features and validation errors of the classifiers obtained from two randomly selected modeling subsets.

Given such instability one cannot reliably draw any conclusions from any single run. For example, if a modeling data set separates the training data well it might be due to over-fitting. Conversely if it separates poorly it might be due to under-fitting. This problem is overcome with our Pseudo Aggregate SVM-RFE method.

**Stability of Pseudo Aggregate SVM-RFE.** Our pseudo aggregate SVM-RFE approach repeats SVM-RFE 100 times using randomized data sets. The data set from each repeat is used to select a set of features, from which a classifier is being built. For simplicity we shall refer to the data set, features and the resulting classifier together from a repeat as a *model*. To counter random fluctuations we consider important features to be those that appear frequently among the 100 classifiers. This reduced the instability caused by randomness. In fact, our belief is as follows: if the two classes are well separated, there should exist a set of features that help to build a good classifier. Most modeling subsets should be able to select these features out and only a limited number of modeling sets might be singular and miss them. Conversely, if the two classes cannot be well separated, no consistently discriminative features exist. Different modeling set may lead to totaly different feature subset. As a result, no feature appears with high frequency in all 100 models. This philosophy, however, is only partially true. When the two classes cannot be separated, the modeling process sometimes can overfit the data by selecting a lot of variables which results in high absolute frequencies for some less important or irrelevant features. Such a phenomenon is usually accompanied by large number of variables and low validation accuracy. To improve the process we propose a more appropriate metric, which we call *relative frequency*. In relative frequency we weight the frequency by two criteria. In the first criteria a variable appearing in short models is weighted more than the variables appearing in long models. This leads to a weight of $h(n_j)$ for a variable in the $j$-th model, with $n_j$ being the number of variables in the $j$-th model. In the second criteria a variable in a model with high predictive accuracy is weighted more than a variable with poor predictive accuracy. This provides another weight $g(A_j)$ for a variable in the $j$-th model, where $A_j$ denotes the accuracy of the $j$-th model computed from the validation process. Mathematically the relative frequency for a variable $x_i$ in a test run of $M$ repeats is defined as

$$(3.1) \qquad rf(x_i) = \frac{1}{M} \sum_{j=1}^{M} g(A_j) h(n_j) \mathbf{1}(x_i \text{ appears in model } j).$$

In our study we always set $M = 100$. Also, we set $g(A_j) = \exp(\frac{A_j - 1}{[2A_j - 1]_+})$ where $[t]_+ = \max\{0, t\}$ and $h(n_j) = [1 - cn_j]_+$ for some constant $c$. For $g(A_j)$ the idea is that if the weight should decay fast if the accuracy is close to 50% or less because it indicates that the classifier is simply not effective. For $h(n_j)$ we put in a penalty for the number of variables used in a model. In our experiments we have chosen $c = 1/30$, which seems to work well.

Our experiments show that features yielded from relative frequency rankings are very stable and consistent. We have performed runs of 100 repeats using different random seeds in MATLAB, and the results are always similar. An additional benefit of using relative frequency instead of absolute frequency is that the existence of an effective classifier is typically accompanied by high relative frequencies for the top features, while low relative frequencies for the top features usually imply poor separability. Hence we can use relative frequency as a simple guide on the separability of the samples. We will show some examples in the next section.

**Results and conclusion.** In Experiment 1 we have performed a run of 100 repeats on the entire Cheng-Gao version of *Dream of the Red Chamber*. Altogether 70 features have appeared in at least one model. However, of those only a small number of them have appeared with high enough frequency to be viewed as being important. We apply cross validation to select the number of features, and the mean cross validation error rate against different number of features is plotted in Figure 1 (a). The figure tells us that 10 to 50 features are enough to tell the style difference between the two parts. Using less characters and words is insufficient, while using more degrades the performance also by bringing in too much noise. The small cross validation error rate is encouraging, and it is already hinting a strong possibility that the two training sample sets have significant stylistic differences to support the two-author hypothesis.

To settle the two-author hypothesis more definitively we apply our classifier on the test data, which until now has never been used during the feature selection and classifier modeling process. In particular we investigate the existence of a chrono-divide in the values obtained through classifier. Figure 1 (b), which plots these values, clearly shows a chrono-divide at Chapter 80: For Chapter 81-90 the classifier yields all negative values while for Chapters 61-80 the classifier yields all positive values with the exception of Chapter 67. Allowing some statistical abberations to occur, our results provide an extremely convincing if not irrefutable evidence that there exist clear stylometric differences between the writings of the first 80 chapters and the last 40 chapters. This difference strongly supports the two-author hypothesis for *Dream of the Red Chamber*. We also note that our investigation did not need to assume that the knowledge that the stylistic change should be at Chapter 80. The fact that the chrono-divide we have detected is indeed at Chapter 80 lends even stronger support to the two-author hypothesis.

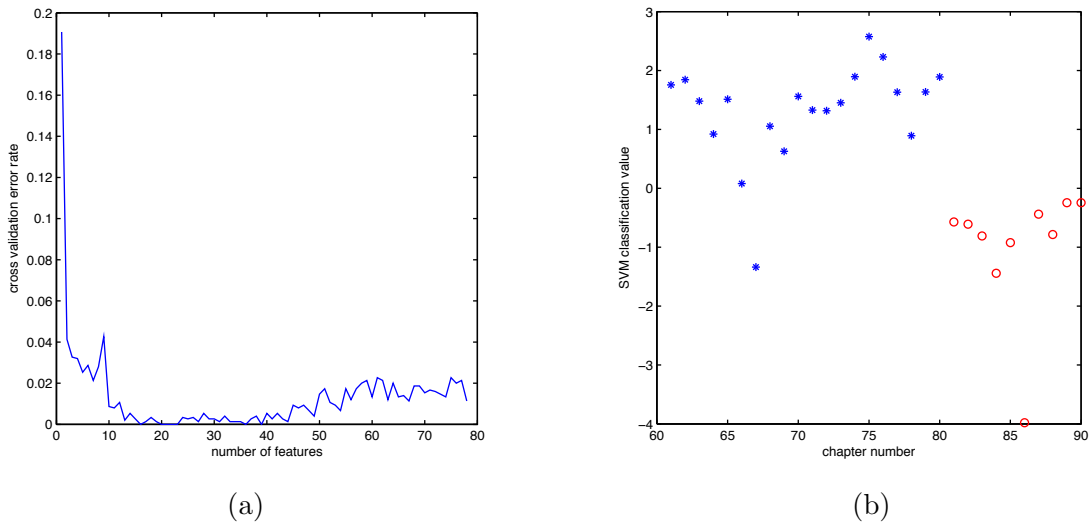

Figure 1. Experiment 1: (a) Mean cross validation error rate; (b) Values of SVM classifier on chapters 60-90.

Interestingly, the fact that Chapter 67 appeared as an "outlier" in our classification serves as further evidence to the validity of our analysis. It was only after the tests we realized that the authorship of Chapter 67 itself is one of the controversies in Redology. Unlike the main controversy about the authorship of the first 80 chapters and the last 40 chapters, experts are less unified in their positions here. Again, our results strongly suggests that Chapter 67 is stylistically different from the rest of the first 80 chapters, and it may not be written by Cao. Our finding is consistent with the conclusion of [5].

3.2. **Non-separability of the first 80 chapters.** To further validate our method we apply the same tests to the first 80 chapters of *Dream of the Red Chamber* to see whether we can get a chrono-divide (Experiment 2). We use the first 30 and last 30 chapters as the training data and leave chapters 31-50 as the test data. Figure 2 shows the mean cross validation error and the values of SVM classifier on the test data chapters 31-50. The experiment shows many more features have been selected in the 100 repeats, implying the difficulty of find a consistent subset of discriminative features. The large errors on the training data also indicate the difficulty for separation. When the classifier is applied to the test data, there is clearly no chrono-divide. This suggests that our method yields a conclusion that is completely consistent with what is known.

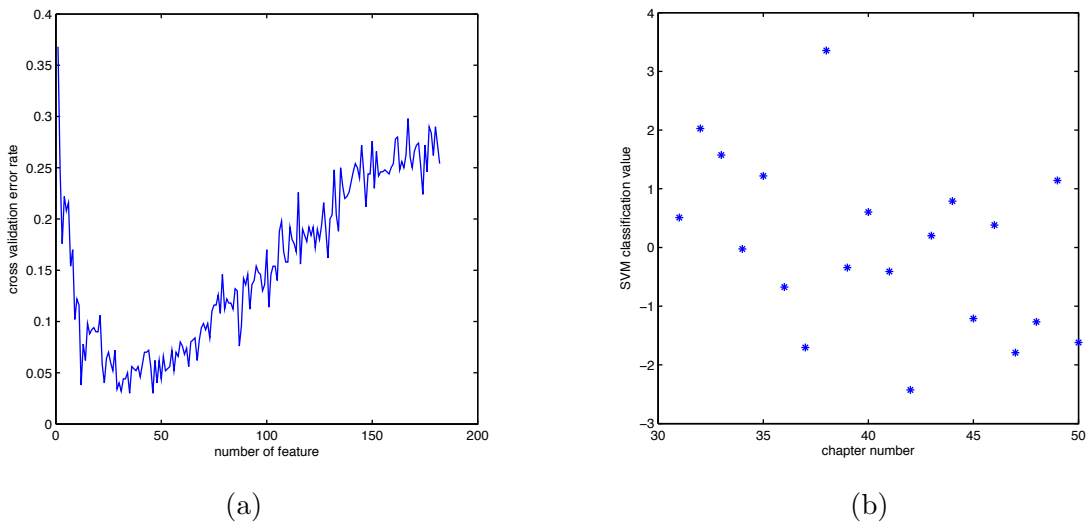

Figure 2. Experiment 2: (a) Mean cross validation error rate; (b) Values of SVM classifier on chapters 31-50. Note there is no chrono-divide.

3.3. **Analysis of chapters 81-120: style change over time.** We next apply our method to the last 40 chapters (Experiment 3). Our first experiment has already confirmed that they are unlikely to be written by Cao. However, there are still debates on whether these were written entirely by one author (most likely Gao himself), or by more than one author. Our mathematical analysis may offer some insight here.

We split the 40 chapters into two subsets as before. The training data include Chapters 81-95 as one class and Chapters 106-120 as another. The test data are the middle 10 chapters. Because of the relatively small number of samples we have subdivided each chapter into 2 sections to increase the sample size. As a result we now have 60 samples in the training data and 20 in test data, with 2 samples corresponding to one chapter. The mean cross validation error of the final classifier and its classification values on the test samples are shown in Figures 3 (a) and (b) respectively.

In this experiment we observe that the performance in terms of both the classifier and feature ranking is noticeably worse than that in Experiment 1 but substantially better than that in Experiment 2. Furthermore, unlike the results from the first two experiments, the values from the classifier show an interesting trend. Compared with Figure 2 (b) where the values appeared to lack any order, the values here exhibit a clear gradual downward shift. On the other hand, compared to Figure 1 (b) the values plotted in 3 (b) do not show a clear sharp chrono-divide, even though the values change gradually from being positive to being negative. What it tells us is that the writing style of the last 80 chapters had undergone a graduate change, but this change is unlikely to be due to change of authorship.

Our results here could be subject to several interpretations. One plausible interpretation is that Gao might indeed obtained some incomplete set of manuscripts by Cao, and tried to complete the novel based on what he had obtained. The style change is a result of the lack of genuine work by Cao as the story developed. A more plausible interpretation is that the last 40 chapters were written by someone such as Gao trying to imitate Cao's style, and over time the author became sloppier and returned more and more to his own style.

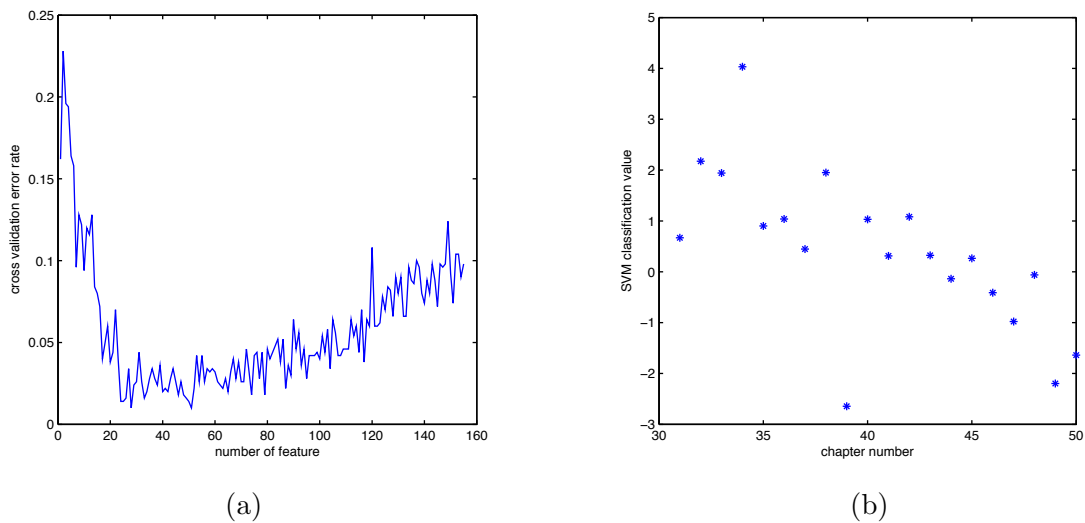

Figure 3. Experiment 3: (a) Mean cross validation error rate; (b) Values of SVM classifier on chapters 96-105, which correspond to the samples 31-50 in all 80 samples. Note two samples come from one chapter in this experiment.

## 4. Analysis of the other three Great Classical Novels

To further bolster the credibility of our approach we test our method on the other three Great Classical Novels in Chinese literature, *Romance of the Three Kingdoms* (三国演义), *Water Margin* (水浒传), and *Journey to the West* (西游记). Unlike *Dread of the Red Chamber*, there is no authorship controversy for these other three novels. Thus if our method is indeed robust we should expect negative answers for the two-author hypotheses for all of them by finding no chrono-divides.

As with *Dream of the Red Chamber*, we split each of the three novels into training samples and test samples. Both *Romance of the Three Kingdoms* and *Water Margin* have 120 chapters. In both cases we designate the first 30 chapters and the last 30 chapters as the two classes of training data, and the middle 60 chapters as test data. For *Journey to the West* the two classes of training data are the first and last 25 chapters respectively, with the middle 50 chapters as test data.

We use the same procedure to test for chrono-divides on the three novels. Compared to *Dream of the Red Chamber*, the selected features show much lower relative frequencies, indicating difficulty in differentiating between the writing styles. Table 2 show the relative frequencies (with $c = 1/30$) of the top 8 features for each of the four Great Classical Novels. Also of note is that in the case of

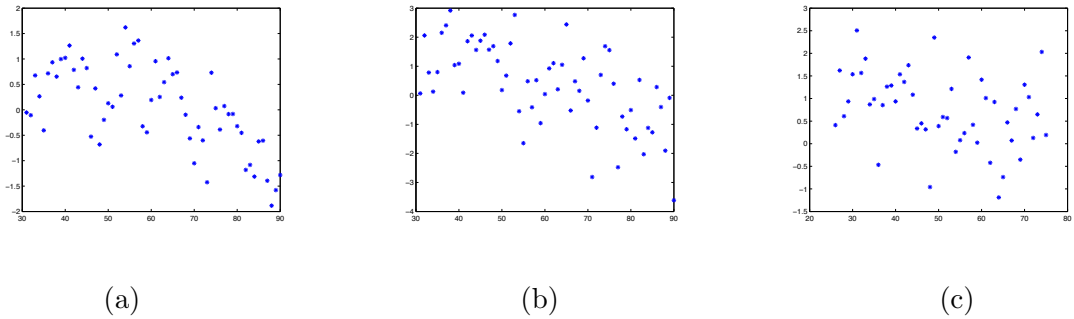

(a)             (b)             (c)

Figure 4. Classification results from the test sampels of the other three classical novels: (a) *Romance of the Three Kingdoms*; (b) *Water Margin*; (c) *Journey to the West*.

*Water Margin*, 51 features are used to build a classifier from the 60 training data, which is clearly another strong indication of the difficulty.

| Novel | Relative frequencies of top 8 features |
|---|---|
| *Dream of the Red Chamber* | 0.57  0.46  0.43  0.36  0.31  0.30  0.29  0.19 |
| *Romance of the Three Kingdoms* | 0.31  0.27  0.26  0.25  0.23  0.22  0.17  0.15 |
| *Water Margin* | 0.18  0.17  0.16  0.16  0.14  0.11  0.11  0.10 |
| *Journey to the West* | 0.03  0.03  0.02  0.02  0.02  0.02  0.02  0.02 |

Table 2. Relative frequencies of the top ranked 8 features in each of the four Great Classical Novels.

Figure 4 plots the values from the classifiers for all three novels. In all cases the values fluctuate in such a way that it is quite clear that no chrono-divides exist, as expected.

This analysis shows that our approach can reliably reject the two-author hypothesis when it is false, lending further support to the effectiveness and robustness of our method.

## 5. Conclusions

Inspired by authorship controversy of *Dream of the Red Chamber* and the application of SVM in the study of literary stylometry, we have developed a mathematically rigorous new method for the analysis of authorship by testing for a chrono-divide in writing styles. We have shown that the method is highly effective and robust. Applying our method to the Cheng-Gao version of *Dream of the Red Chamber* has led to convincing if not irrefutable evidence that the first 80 chapters and the last 40 chapters of the book were written by two different authors. Furthermore, our analysis has unexpectedly provided strong support to the hypothesis that Chapter 67 was not the work of Cao Xueqin either.

The methodology in this paper is rather effective in selecting the most important features for classification through a new ranking system based on relative frequency. A series of future experiments should be included in the application of this methodology to wider range of works.

It is worth mentioning that there are several other attempts to complete *Dream of the Red Chamber* from the its first 80 chapters, among them is *Continued Dream of the Red Chamber* (续红楼梦) by Qi Zichen (秦子忱). Using the same features for building the classifier in Experiment 1, we can compute the Euclidean distances between all chapters and their distances of chapters from *Continued Dream of the Red Chamber*, see Figure 5. Surprisingly, although these features are obtained in favor of the differences between Cao and Cheng-Gao, they lead to even larger distance between the first 80 chapters and those chapters of *Continued Dream of the Red Chamber*. It obviously implies that the style of the 40 chapters by Cheng-Gao are more similar to the 80 chapters by Cao compared to *Continued dream of the Red Chamber*. Maybe that's why the Cheng-Gao version is more popular than other versions.

**Acknowledgement.** Throughout this project, Haiyi Jiang has enthusiastically participated in the discussions on the project, and his knowledge of *Dream of the Red Chamber* has been valuable to the study. The authors would like to thank him in particular for his support.

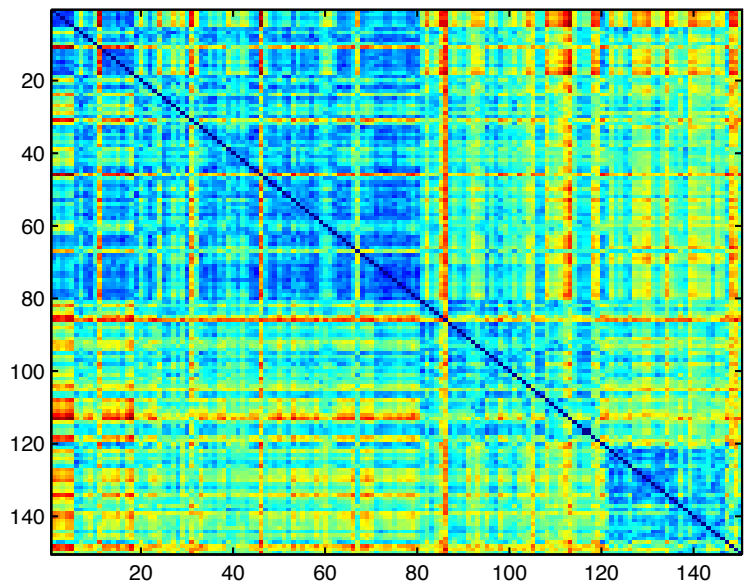

Figure 5. Distances between the first 80 chapters of the Cheng-Gao version, the last 40 chapters of the Cheng-Gao version, and 30 chapters of *Continued Dream of the Red Chamber* .

Department of Mathematics, Michigan State University, East Lanisng, MI 48824, USA.

*E-mail address*: `hxf0204@gmail.com`

Department of Mathematics, Michigan State University, East Lanisng, MI 48824, USA.

*E-mail address*: `ywang@math.msu.edu`

Department of Mathematical Sciences, Middle Tennessee State University, Murfreesboro, TN 37132, USA.

*E-mail address*: `qwu@mtsu.edu`